\documentclass[10pt]{article}

\usepackage[utf8]{inputenc} 
\usepackage[T1]{fontenc}    
\usepackage{hyperref}       
\usepackage{url}            
\usepackage{booktabs}       
\usepackage{amsfonts}       
\usepackage{nicefrac}       
\usepackage{microtype}      

\usepackage{algorithmicx,algorithm}
\usepackage{amsmath, amsthm, amssymb, multirow, paralist}
\usepackage[numbers]{natbib}

\usepackage[noend]{algpseudocode}

\usepackage{graphicx}

\usepackage{comment}
\usepackage{color}
\usepackage{dsfont}

\usepackage{wrapfig}

\usepackage[body={6.5in,8in}]{geometry}

\newtheorem{thm}{Theorem}

\newtheorem{lemma}{Lemma}

\newcommand{\argmax}{\mathop{\arg\max}}

\def \R {\mathbb{R}}

\def \X {\mathbf{X}}
\def \Id {\mathds{1}}

\def \Z {\mathbf{Z}}

\def \Y {\mathbf{Y}}
\def \E {\mathbb{E}}
\def \P {\mathbb{P}}
\def \M {\mathbf{M}}
\def \hM {\widehat{M}}
\def \proj {\mathcal{R}}
\def \hX {\widehat{X}}
\def \hZ {\widehat{Z}}
\def \L {\mathcal{L}}
\def \bL {\bar{\mathcal{L}}}
\def \bhM {\mathbf{\hM}}
\def \A {\mathbf{A}}
\def \I {\mathbf{I}}
\def \U {\mathbf{U}}
\def \V {\mathbf{V}}
\def \G {\mathbf{G}}
\def \bL {\mathbf{L}}
\def \bR {\mathbf{R}}
\def \B {\mathbf{B}}
\def \bhX {\mathbf{\hX}}
\def \bhZ {\mathbf{\hZ}}

\title{Matrix Co-completion for Multi-label Classification with Missing
Features and Labels}
\author{
  Miao Xu$^1$, Gang Niu$^{1}$, Bo Han$^{2,1}$, Ivor W. Tsang$^2$, Zhi-Hua Zhou$^3$, Masashi Sugiyama$^{1,4}$\\
  $^1$RIKEN Center for Advanced Intelligence Project, Tokyo, Japan\\
  $^2$Center for Artificial Intelligence, University of Technology Sydney, Sydney, Australia\\
 $^3$Nanjing University, Nanjing, China\\
 $^4$University of Tokyo, Tokyo, Japan\\
  \texttt{\{miao.xu, gang.niu, bo.han@riken.jp\}, ivor.tsang@uts.edu.au},\\
  \texttt{zhouzh@lamda.nju.edu.cn, sugi@k.u-tokyo.ac.jp}
}

\begin{document}
\maketitle

\begin{abstract}
We consider a challenging multi-label classification problem
where both feature matrix $\X$ and label matrix $\Y$ have missing entries.
An existing method concatenated $\X$ and $\Y$ as $[\X; \Y]$ and
applied a matrix completion (MC) method to fill the missing entries,
under the assumption that $[\X; \Y]$ is of low-rank.
However, since entries of $\Y$ take binary values in the multi-label setting,
it is unlikely that $\Y$ is of low-rank. Moreover, such assumption implies a linear relationship between $\X$ and $\Y$ which may not hold in practice.
In this paper, we consider a \emph{latent} matrix $\Z$ that produces
the probability $\sigma(Z_{ij})$ of generating label $Y_{ij}$, where $\sigma(\cdot)$ is nonlinear. Considering label correlation, we assume $[\X; \Z]$ is of low-rank, and propose an MC algorithm based on subgradient descent
named \emph{co-completion (COCO)} motivated by elastic net and one-bit MC. We give a theoretical bound on the recovery effect of COCO
and demonstrate its practical usefulness through experiments.
\end{abstract}
%

\section{Introduction}
\emph{Multi-label learning}~\cite{DBLP:reference/ml/ZhouZ17}, which allows an instance to be associated with multiple labels simultaneously, has been applied successfully to various real-world problems, including images~\cite{DBLP:conf/icml/ChenZW13}, texts~\cite{DBLP:conf/nips/NguyenBRC14} and biological data~\cite{DBLP:conf/aaai/ChenCWZ15}. An important issue with multi-label learning is that collecting all labels requires investigation of a large number of candidate labels one by one, and thus labels are usually missing in practice due to limited resources.

Multi-label learning with such missing labels, which is often called \emph{weakly supervised} multi-label learning (WSML), has been investigated thoroughly~\cite{DBLP:conf/icml/Yu0KD14,DBLP:conf/aaai/SunZZ10,DBLP:conf/cvpr/BucakJJ11}. Among them, the most popular methods are based on \emph{matrix completion (MC)}~\cite{DBLP:conf/icml/Yu0KD14,DBLP:conf/nips/GoldbergZRXN10,DBLP:conf/nips/XuJZ13},
which is a technique to complete an (approximately) low-rank matrix with uniformed randomly missing entries~\cite{DBLP:journals/focm/CandesR09,DBLP:journals/pieee/CandesP10}.

The MC-based WSML methods mentioned above assume that only the label matrix $\Y$ has missing entries, while the feature matrix $\X$ is complete. However, in reality, features can also be missing~\cite{DBLP:conf/icml/DekelS08}. To deal with such missing features, a naive solution is to first complete the feature matrix $\X$ using a classical MC technique, and then employ a WSML method to fill the label matrix $\Y$. However, such a two-step approach may not work well since recovery of $\X$ is performed in an unsupervised way (i.e., label information $\Y$ is completely ignored). Thus, when facing a situation with both label and feature missing, it would be desirable to employ the label information $\Y$ to complete the feature matrix $\X$ in a supervised way.
Following this spirit, \cite{DBLP:conf/nips/GoldbergZRXN10} proposed concatenating
the feature matrix $\X$ and label matrix $\Y$ into a single big matrix $[\X;\Y]$,
and employed an MC algorithm to recover both features and labels simultaneously.

\begin{wrapfigure}{r}{0.4\textwidth}
\centering
\includegraphics[width=0.38\textwidth]{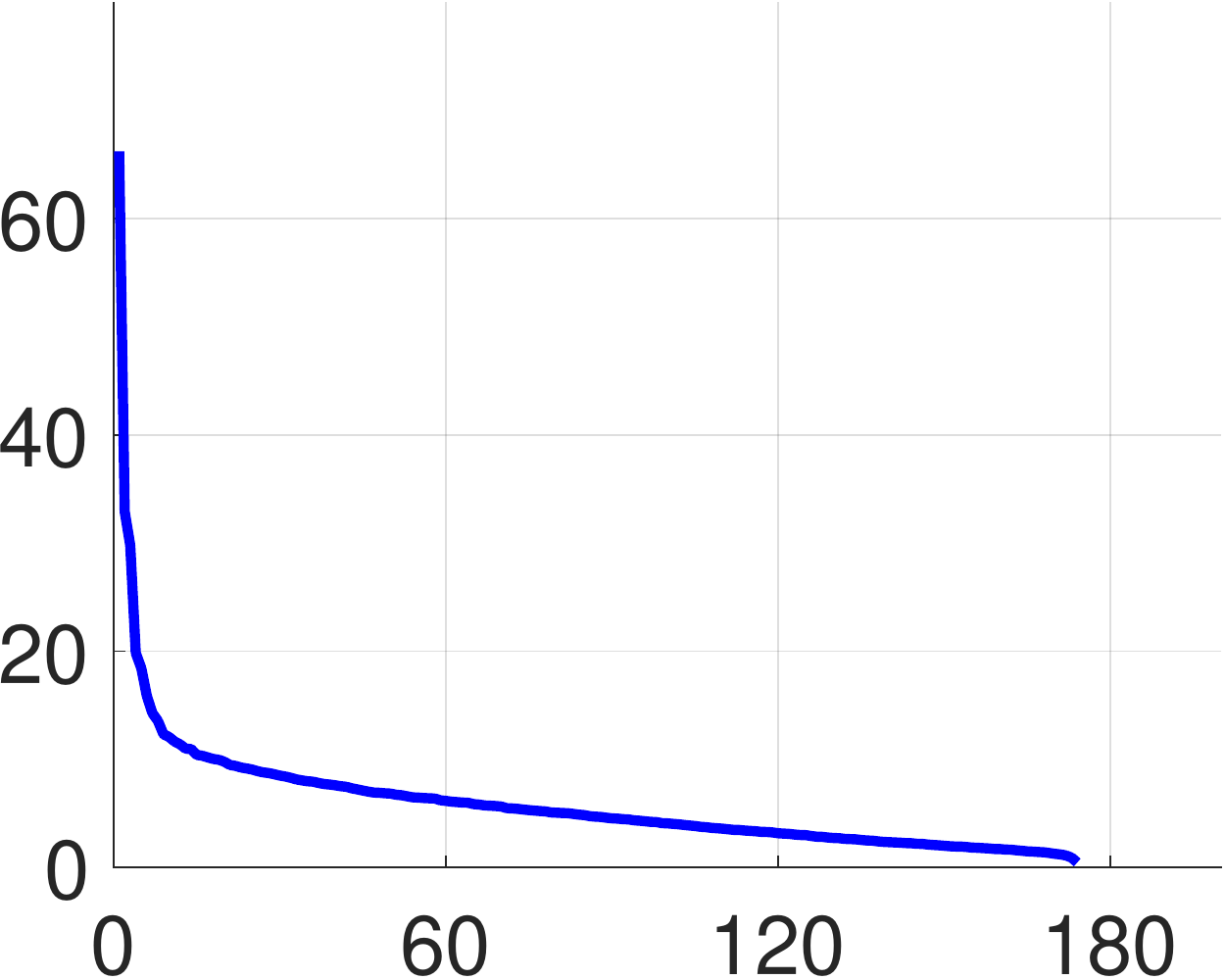}
\caption{$174$ singular values of the CAL500 data sorted in descending order. }
\label{illustration-general}
\end{wrapfigure}

In the WSML methods reviewed above,
the label matrix $\Y$ is commonly assumed to be of (approximately) low-rank,
based on the natural observation that labels are correlated in the multi-label setting.
However, such a low-rank assumption on $\Y$ may not be true in reality
since entries of $\Y$ take binary values and thus $\Y$ is unlikely to be of low-rank.
Indeed, as we observe in Figure~\ref{illustration-general},
the singular values of the label matrix $\Y$ of the CAL500 data~\cite{DBLP:journals/taslp/TurnbullBTL08}
have a heavy tail and thus such a low-rank assumption on $\Y$ may not be reasonable.
Another assumption of~\cite{DBLP:conf/nips/GoldbergZRXN10} is that there is a linear relationship between the feature and label matrices. Such an assumption may not hold well on some datasets, where classical multi-label methods learn a nonlinear classifier~\cite{DBLP:conf/nips/ElisseeffW01,Zhang:2006:MNN:1159162.1159294}.

In this paper, we propose a method to deal with WSML learning when both features and annotations are incomplete. Motivated by~\cite{DBLP:conf/aaai/HuangGZ14} learning a low-dimensional shared subspace between labels and features, we assume that there is some latent matrix $\Z$ generating the annotation matrix. More specifically, each entry of $\Z$ will be mapped by a nonlinear function $\sigma(\cdot)$ to $[0,1]$, which corresponds to the probability of setting the corresponding entry of $\Y$ to be $1$. Considering the label correlation, we assume $[\X;\Z]$ is low-rank. Motivated by elastic net~\cite{Zou05regularizationand} and one-bit MC~\cite{DBLP:journals/corr/abs-1209-3672,NIPS2016_6567,DBLP:conf/aistats/NiG16}, we propose a subgradient-based MC method named \emph{co-completion (COCO)} which can recover $\X$ and $\Y$ simultaneously. Furthermore, we give a theoretical bound on the recovery effect of COCO. In the experiments, we demonstrate that COCO not only has a better recovery performance than baseline, but also achieves better test error when the recovered data are used to train new classifiers.

The following of the paper is organized as follows. In Section~\ref{sec:alg} we introduce our proposed algorithm COCO, followed by theoretical guarantee in Section~\ref{sec:theory} and experimental results in Section~\ref{sec:experiments}. We finally give conclusion and future work in Section~\ref{sec:con}.

%
%
%

\section{Algorithm}~\label{sec:alg}
In this section, we will first give the formal definition of our studied problem. Then, we will give our learning objective, as well as the optimization algorithm.

\subsection{Formulation}
We will assume $\X\in\R^{n\times d}$ is the feature matrix, in which $n$ is the number of instances and $d$ is the number of features. There is a label matrix $\Y\in\{0,1\}^{n\times l}$ where $l$ is the number of labels in multi-label learning. We assume that $\Z\in\R^{n\times l}$ generates the label matrix $\Y$, that is, $\P[Y_{i,j}=1]=\sigma(Z_{i,j})$. In the following, we will assume $\sigma(\cdot)$ is the sigmoid function, i.e., $\P[Y_{i,j}=1]=1/(1+\exp(-Z_{i,j}))$.

Our basic assumption is that $\X$ and $\Z$ are concatenated together into a big matrix $\M$ which is written as $[\X;\Z]$, and $\M$ is low-rank. Note that previous work~\cite{DBLP:conf/nips/GoldbergZRXN10} has two assumptions when using matrix completion to solve a multi-label problem. One is that the label matrix $\Y$ has a linear relationship with the feature matrix $\X$. However, such an assumption may not hold well on real data, otherwise, there may not exist so many algorithms learning a nonlinear mapping between features and labels. Another assumption is that the label matrix $\Y$ is low-rank. Such an assumption is motivated by the fact that labels in multi-label learning are correlated, thus only a few factors determine their values. However, we want to argue that a sparse $0/1$ matrix may not be low-rank, and instead, we assume that the latent matrix $\Z$ generating labels is low-rank. Thus in our problem, we assume the concatenate of $\X$ and $\Z$ forms a low-rank matrix, and will recover such a matrix when entries in both $\X$ and $\Z$ are missing.

Similarly to previous works~\cite{DBLP:conf/icml/Yu0KD14,DBLP:conf/nips/XuJZ13}, we assume that data are uniformly randomly missing with probability $1-|\Omega|/n(d+l)$, where, for $[n]=\{1,\ldots,n\}$, $\Omega\subset [n]\times [d+l]$ for  contains all the indices of observed entries in the matrix $\M$. Let $\Omega_X, \Omega_Y\subset \Omega$ be the subsets containing all the indices of observed entries in $\X$ and $\Y$ respectively. Based on all these notations, in the following we will give our learning objective.

\subsection{Learning Objective}

In our learning objective, we need to consider three factors. One focuses on the feature matrix. To recover the feature matrix, a classical way is to use the Frobenius norm on those observed entries, i.e.,
\begin{eqnarray*}
\|\proj_{\Omega_X}(\widehat \X-\X)\|_\mathrm{F}^2,
\end{eqnarray*}
where
\begin{eqnarray*}
[\proj_{\Omega_X}(\A)]_{i,j}=\begin{cases}
A_{i,j}& $if$ \ (i,j)\in \Omega_X,\\
0 & $otherwise.$
\end{cases}
\end{eqnarray*}

Note that the Frobenius norm on matrices is corresponding to the L2 norm on vectors, and the trace norm on matrix singular values is similar to the L1 norm on vectors. Motivated by the advantage of the elastic net~\cite{Zou05regularizationand} which uses both the L1 norm and L2 norm for regularization, we additionally consider optimizing the trace norm of the difference between the recovered feature matrix and the observed feature matrix, i.e.,
\begin{eqnarray*}
\|\proj_{\Omega_X}(\widehat \X-\X)\|_\mathrm{tr}.
\end{eqnarray*}

For the label matrix $\Y$, motivated by previous work on one-bit matrix completion~\cite{DBLP:journals/corr/abs-1209-3672}, we will consider the log-likelihood of those observed entries, i.e.,
\begin{eqnarray*}
\sum_{(i,j)\in\Omega_Y} [\Id_{Y_{i,j-d}}=1] \log\sigma(Z_{ij})+ [\Id_{Y_{i,j-d}}=0] \log (1-\sigma(Z_{ij})),
\end{eqnarray*}
where $\Id$ is the indicator function. Note that we will minimize the negative log-likelihood instead of maximizing the log-likelihood, in order to agree with other components in the objective.

To incorporate all the above conditions into consideration, we have our final learning objective,
\begin{eqnarray}\nonumber
\min_{\widehat\X,\widehat\Z} && -\sum_{(i,j)\in\Omega_Y} [\Id_{Y_{i,j-d}}=1] \log\sigma(\hat Z_{ij})+ [\Id_{Y_{i,j-d}}=0] \log (1-\sigma(\hat Z_{ij}))\\\label{eq:mainopt}
&&+\lambda_1 \|\widehat \M\|_\mathrm{tr}+\|\proj_{\Omega_X}(\widehat \X-\X)\|_\mathrm{F}^2+\lambda_2 \|\proj_{\Omega_X}(\widehat \X-\X)\|_\mathrm{tr},
\end{eqnarray}
where $\widehat \M$ is the concatenation of $\widehat \X$ and $\widehat \Z$, i.e., $[\widehat\X;\widehat\Z]$.

\subsection{Optimization}

Previous deterministic algorithms on trace norm minimization~\cite{RePEc:cor:louvco:2007076,paul-tseng} always assume that the loss function is composed of two parts. One part is a differential convex function, and another part is the trace norm on the whole matrix, which is not differentiable but convex. In this way, it is easy for them to find a closed-form solution using algorithms such as proximal gradient descend, because minimizing the trace norm on the whole matrix plus a simple loss function will have a closed-form solution~\cite{RePEc:cor:louvco:2007076}.

However, in our problem Eq~(\ref{eq:mainopt}), besides the simple trace norm on the whole matrix, we still have the trace norm on the submatrix, while $\proj_{\Omega_X}(\widehat \X-\X)$ is a linear transformation of the whole matrix $\widehat \M$. Thus classical methods based on proximal gradient descent cannot be employed.


We divide the learning objective into two parts, and consider each part seperatedly. One part is
\begin{eqnarray}\label{eq:subp}
f(\mathbf{\hM})+\lambda_2h(\mathbf{\hM}),
\end{eqnarray}
where
\begin{eqnarray*}
f(\mathbf{\hM})=-\sum_{(i,j)\in\Omega_\Z}\left[[\Id_{Y_{i,j-d}}=1]\log\sigma(\hM_{i,j})+[\Id_{Y_{i,j-d}}=0]\log(1-\sigma(\hM_{i,j}))\right]\\+\sum_{(i,j)\in\Omega_\X} (\hM_{i,j}-M_{i,j})^2,
\end{eqnarray*}
and
\begin{eqnarray*}
h(\mathbf{\hM})=\lambda_2 \|\proj_{\Omega_\X}(\mathbf \hM\I_{(d+l)\times d}-\M\I_{(d+l)\times d})\|_\mathrm{tr},
\end{eqnarray*}
where $\I_{(d+l)\times d}$ is a matrix whose diagonal entries are $1$ and other entries are $0$.

Another part contains only $\|\widehat \M\|_\mathrm{tr}$. Note that in previous works on the stochastic L1 loss minimization problem~\cite{DBLP:conf/icml/Shalev-ShwartzT09,DBLP:conf/nips/LangfordLZ08}, they first perform gradient descent on the loss function without considering the L1 loss part, and then derived a closed-form solution for the L1 loss part. Motivated by this, we will first perform gradient descent on~Eq~(\ref{eq:subp}) and then obtain a closed-form solution taking $\|\widehat\M\|_\mathrm{tr}$ into consideration.

$f(\mathbf{\hM})$ is convex and it is easy to calculate the derivation. To calculate the subgradient of $h(\mathbf{\hM})$, we will need the following results:
\begin{lemma}
(Subgradient of the trace norm~\cite{articlew1992}) Let $\X\in\R^{m\times n}$ with $m\ge n$, and let $\X=\U\mathbf{\Sigma}\V$ be an singular value decomposition (SVD) of $\X$. Let $r=\mathrm{rank}(\X)$. Then,
\begin{eqnarray*}
\U_{1:m,1:r}\V^\top_{1:n,1:r}\in \|\X\|_\mathrm{tr}.
\end{eqnarray*}
\end{lemma}

In this way,the subgradient of $h(\mathbf{\hM})$ is given by
\begin{eqnarray*}
\proj_{\Omega_\X}(U_hV_h^\top) \I_{(d+l)\times d}^\top,
\end{eqnarray*}
where $\mbox{svd}(\proj_{\Omega_\X}(\mathbf \hM\I_{(d+l)\times d}-\M\I_{(d+l)\times d}))=U_h\Sigma_h V_h^\top$.

We will perform iterative optimization. In the $t$th iteration, after we have the subgradient $\G_t$ of Eq~(\ref{eq:subp}), $\widehat{M}$ will be updated using gradient descent by
\begin{eqnarray}\label{eq:update1}
\widetilde\M_t=\widehat \M_{t-1}-\eta\G_t,
\end{eqnarray}
where $\eta>0$ is the step size. We then have a closed-form solution of $\widehat \M_t$ taking the trace norm into consideration, which is,
\begin{eqnarray}\label{eq:update2}
\mathcal{D}_{\lambda_1}[\widetilde\M_t]=\U_t\mathcal{D}_{\lambda_1}[\mathbf{\Sigma}_t]\V_t^\top,
\end{eqnarray}
where $\U_t\mathbf{\Sigma}_t\V_t^\top$ is the SVD of $\widetilde\M_t$ and
\begin{eqnarray*}
\mathcal{D}_{\lambda_1}[\mathbf{\Sigma}_t]=\text{diag}[\max(0,\sigma_1-\lambda),\ldots,\max(0,\sigma_n-\lambda)].
\end{eqnarray*}
We will call our proposed method \emph{co-completion (COCO)} and give the whole process in Algorithm~\ref{alg:coco}.

Note that such a solution coincides with works on stochastic trace norm minimization~\cite{DBLP:conf/icml/AvronKKS12,DBLP:journals/corr/0005YJZ15b}. In both works, they constructed a random probe matrix, and multiplied the gradient with the probe matrix in each iteration to generate a stochastic gradient. In this way, the expectation of the stochastic gradient calculated in each iteration will be the exact gradient, which agrees with the principle of stochastic gradients in ordinary stochastic gradient descent (SGD). ~\cite{DBLP:journals/corr/0005YJZ15b} provided a theoretical guarantee of the $O(\log T/\sqrt{T})$ convergence rate for such a kind of problems. As their objective is to save space for trace norm minimization, here we will not consider the space limitation problem, and will use plain gradient descent instead of subgradient descent. However, their convergence results on SGD can be used as a weak guarantee for the convergence of our algorithm.

\begin{algorithm}[!t]
	\centering
	\caption{COCO}
	\label{alg:coco}
	\begin{algorithmic}[1]
	\State Input the number of trials $T$, the step size $\eta$ and $\lambda_1$, $\lambda_2$;
		\State Initialize $\widehat \M_0$;
		\For{$t=1,2,\ldots,T$};
	    \State Update $\widetilde\M_t$ using Eq~(\ref{eq:update1});
	    \State Calculate $\widehat \M_t$ using Eq(\ref{eq:update2});
	    \EndFor
	    \State Output $\widehat \M_{T}$
	\end{algorithmic}
\end{algorithm}

\section{Theory}~\label{sec:theory}
In this section, we give a bound on the following optimization problem:
\begin{eqnarray}\nonumber
\max_{\mathbf{\hM}}& \sum_{(i,j)\in\Omega_\Z}[\Id_{Y_{i,j-d}}=1]\log\sigma(\hM_{i,j})+[\Id_{Y_{i,j-d}}=0]\log(1-\sigma(\hM_{i,j}))& \\\nonumber
&\;\;\;\;\;\;\;\;\;\;\;\;\;\;\;\;\;\;\;\;\;\;\;\;\;\;\;\;\;\;\;\;\;\;\;\;\;\;\;\;\;\;\;\;\;\;\;\;\;\;\;\;\;\;\;\;\;\;\;\;\;\;\;\;\;\;\;\;\;\;-\sum_{(i,j)\in\Omega_\X} (\hM_{i,j}-M_{i,j})^2&\\\nonumber
\text{s.t.}& \|\mathbf{\hM}\|_\mathrm{tr}\le \alpha\sqrt{rn(d+l)},&\\\label{eqn:opt-obj}
&\|\proj_{\Omega_\X}(\mathbf \hX-\X)\|_\mathrm{tr}\le \beta L_\gamma \sqrt{rn(d+l)} \sqrt\frac{l}{d}.&
\end{eqnarray}
Note that if we change the $\max$ operator in Eq~(\ref{eqn:opt-obj}) to $\min$ and change the objective to its additive inverse, we will have an equivalence of Eq~(\ref{eqn:opt-obj}). We can then use Lagrange multiplier and add the two inequality constraints into the objective. In this way, the problem will have similar form as Eq~(\ref{eq:mainopt}). Thus by appropriately setting parameters $\lambda_1$ and $\lambda_2$ in Eq~(\ref{eq:mainopt}), the maximization problem Eq~(\ref{eqn:opt-obj}) and the minimization problem Eq~(\ref{eq:mainopt}) will be equal.

We assume that
\begin{eqnarray*}
\L_{\Y,\Omega}(\mathbf{\hM})=\sum_{(i,j)\in\Omega_\Z}[\Id_{Y_{i,j-d}}=1]\log\sigma(\hM_{i,j})+[\Id_{Y_{i,j-d}}=0]\log(1-\sigma(\hM_{i,j}))\\\;\;\;\;\;\;\;\;\;\;\;\;\;\;\;\;\;\;\;\;\;\;\;\;\;\;\;\;\;\;\;\;\;\;\;\;\;\;\;\;\;\;\;\;\;\;\;\;\;\;\;\;\;\;\;\;\;\;\;\;
 -\sum_{(i,j)\in\Omega_\X} (\hM_{i,j}-M_{i,j})^2
\end{eqnarray*}
and
\begin{eqnarray*}
\bhM^*=\argmax_{\mathbf{\hM}}\L_{\Y,\Omega}=[\bhX^*;\bhZ^*].
\end{eqnarray*}

We further define
\begin{eqnarray*}
\bL_{\Y,\Omega}(\mathbf{\hM})=\L_{\Y,\Omega}(\mathbf{\hM})-\L_{\Y,\Omega}([\X;\mathbf{0}])
\end{eqnarray*}
in which $\mathbf{0}$ is an all-zero matrix of size $n\times l$.

Since $[\X;\mathbf{0}]$ is a constant matrix, minus $\L_{\Y,\Omega}([\X;\mathbf{0}])$ will not affect optimizing of the objective function, i.e., maximizing $\L_{\Y,\Omega}(\mathbf{\hM})$ and $\bL_{\Y,\Omega}(\mathbf{\hM})$ under the same constraints will result in the same $\mathbf{\hM}^*$.

In the following, we will start deriving our theoretical results.

\begin{lemma}\label{lem:main}
Let $G\subset\R^{n\times(d+l)}$ be
\begin{eqnarray*}
G=\left\{\mathbf{\hM}\in\R^{n\times(d+l)},\|\mathbf{\hM}\|_\mathrm{tr}\le \alpha\sqrt{rn(d+l)},\|\proj_{\Omega_\X}(\mathbf \hX-\X)\|_\mathrm{tr}\le \beta L_\gamma \sqrt{rn(d+l)} \sqrt\frac{l}{d}\right\}
\end{eqnarray*}
for some $r\le \min\{d_1,d_2\}$, and $\alpha\ge 0$. Then
\begin{eqnarray*}
&&\P\left[\sup_{\bhM\in G}\left\|\bL_{\Y,\Omega}(\bhM)-\E[\bL_{\Y,\Omega}(\bhM)]\right\|\ge\right.\\
&&\;\;\;\;\;\;\;\;\;\left.\left(C_0\alpha L_\gamma\sqrt{r}\sqrt{|\Omega|(n+d+l)+n(d+l)\log(n+d+l)}\sqrt{\frac{l}{d+l}}\right)\right]\le \frac{C}{n+d+l},
\end{eqnarray*}
where $C_0$ and $C$ are constants, and the expectation are both over the choice of $\Omega$ and the draw of $\Y$.
\end{lemma}

With Lemma~\ref{lem:main}, we can have the following results:
\begin{thm}\label{thm:thm1}
Assume that $\|\M\|_\mathrm{tr}\le \alpha\sqrt{rn(d+l)}$ and the largest entry of $\M$ is less than $\gamma$. Suppose that $\Omega$  is chosen independently at random following a binomial model with probability $|\Omega|/(n(d+l))$. Suppose that $\Y$ is generated using $\sigma(\Z)$. Let $\bhM^*$ be the solution to the optimization problem Eq~(\ref{eqn:opt-obj}). Then with a probability at least $1-C/(n+d+l)$, we have
\begin{eqnarray*}
&&\mathrm{KL}(\sigma(\Z)\|\sigma(\bhZ^*))+\frac{d}{l}\left[\frac{1}{n(d+l)}\|\bhX^*-\X\|^2_\mathrm{F}\right]\\
&\le& 2C_0\alpha L_\gamma\sqrt{\frac{r(n+d+l)}{|\Omega|}}\sqrt{1+\frac{(n+d+l)\log(n+d+l)}{|\Omega|}}\sqrt{\frac{d+l}{l}},
\end{eqnarray*}
\end{thm}
where $\mathrm{KL}$ denotes the Kullback-Leibler on two matrices. For $\A,\B\in\R^{n\times m}$ it is defined as
\begin{eqnarray*}
\mathrm{KL}(\A\|\B)=\frac{1}{nm}\sum_{ij}\mathrm{KL}(A_{ij}\| B_{ij}).
\end{eqnarray*}

By enforcing $\gamma\rightarrow\infty$ and using the fact that that $L_\gamma=1$ when $\sigma(\cdot)$ is a sigmoid function~\cite{DBLP:journals/corr/abs-1209-3672}, we can have our main result:
\begin{thm}\label{thm:thmmain}
Assume that $\|\M\|_\mathrm{tr}\le \alpha\sqrt{rn(d+l)}$. Suppose that $\Omega$  is chosen independently at random following a binomial model with probability $|\Omega|/(n(d+l))$. Suppose that $\Y$ is generated using $\sigma(\Z)$. Let $\bhM^*$ be the solution to the optimization problem Eq.(\ref{eqn:opt-obj}). Then with probability at least $1-C/(n+d+l)$, we have
\begin{eqnarray*}
&&\mathrm{KL}(\sigma(\Z)\|\sigma(\bhZ^*))+\frac{d}{l}\left[\frac{1}{n(d+l)}\|\bhX^*-\X\|^2_\mathrm{F}\right]\\
&\le& 2C_0\alpha \sqrt{\frac{r(n+d+l)}{|\Omega|}}\sqrt{1+\frac{(n+d+l)\log(n+d+l)}{|\Omega|}}\sqrt{\frac{d+l}{l}}.
\end{eqnarray*}
Furthermore, as long as $|\Omega|\ge (n+d+l)\log(n+d+l)$, and further assuming that $r\ll \log(n+d+l)$, we will have
\begin{eqnarray*}
\mathrm{KL}(\sigma(\Z)\|\sigma(\bhZ^*))+\frac{d}{l}\left[\frac{1}{n(d+l)}\|\bhX^*-\X\|_\mathrm{F}^2\right]
\le2\sqrt{2}C_0\alpha \sqrt{\frac{d+l}{l}}.
\end{eqnarray*}
\end{thm}

\paragraph{Remarks} Theorem 2 tells us that the average KL-divergence of the recovered $\sigma(\widehat\Z^*)$ and $\sigma(\Z)$, together with the average Frobenius norm of $\widehat\X^*-\X$ weighted by $d/l$ are bounded above by $O(\sqrt{(d+l)/l})$ if $|\Omega|\ge O(m\log m)$, in which $m=\max(n,d+l)$. Note that when $d\gg l$, the $\X$ part will take the majority of $\M$, and the bound implies that we can have a nearly perfect feature recovery result with sample complexity $O(mlog m)$, agreeing with previous perfect-recovery results although the confidence is degenerated a bit from $1-m^{-\beta}$ where $\beta>1$ to $1-m^{-1}$~\cite{DBLP:journals/jmlr/Recht11}. Otherwise if $l\gg d$, our bound also agrees with previous bound on one-bit matrix completion~\cite{DBLP:journals/corr/abs-1209-3672}.

\section{Experiments}~\label{sec:experiments}
We evaluate the proposed algorithm COCO on both synthetic and real data sets. Our implementation is in Matlab except the neural network which is implemented in Python and used to show the generalization performance of classifiers trained on recovered data.

\subsection{Experimental Results on Synthetic Data}
\label{sec:expsim}
Our goal is to show the recovery effect of our proposed algorithm on both the feature matrix and label matrix. We will also show how adding the term $\|\proj_{\Omega_X}(\widehat \X-\X)\|_\mathrm{tr}$ can enhance our recovery effect.

\paragraph{Settings and Baselines} To create synthetic data, following previous works generating a low-rank matrix~\cite{DBLP:journals/siamjo/CaiCS10}, we first generate a random matrix $\bL\in\R^{n\times r}$ and $\bR\in\R^{r\times (d+l)}$ with each entry drawn uniformly and independently randomly from $[-5,,5]$. We then construct $\M$ by $\bL\cdot \bR$. The first $d$ columns of $\M$ is regarded as the feature matrix $\X$ and the rest is regarded as the $\Z$ matrix. We then set each entry of $\Y\in\R^{n\times l}$ by $1$ with probability $\sigma(Z_{ij})$ and $0$ with $1-\sigma(Z_{ij})$. Here $\sigma(\cdot)$ is the sigmoid function. Finally both $\X$ and $\Y$ are observed with probability $\omega\%$ for each entry.

We set a variety of different numbers to $n$, $d$, $l$, $r$, $\omega$. More specially, $n\in \{1000,5000,10000\}$, $d\in\{500,800\}$, $l\in\{100,300\}$, $r\in\{10,20\}$, $\omega\in\{0.2,0.3,0.4\}$. In the experiments, we weight $\|\proj_{\Omega_X}(\widehat \X-\X)\|_\mathrm{F}^2$ by $0.01$ and set all other weight parameters to be~$1$ . The step size is set initially as $100$ and decays at the rate of $0.99$, i.e., $\eta_t=0.99\eta_{t-1}$, until it is below $40$. We will compare two cases: One is the parameter $\lambda_2=0$ without considering the $\|\proj_{\Omega_X}(\widehat \X-\X)\|_\mathrm{tr}$ term in the optimization; another is $\lambda_2=1$ motivated by the elastic net. The Maxide method is to first complete features using proximal graident descent~\cite{paul-tseng} and then perform weakly supervised multi-label learning~\cite{DBLP:conf/nips/XuJZ13}. The Mc method is to complete the concatenate of $\X$ and $\Y$, which is proposed in~\cite{DBLP:conf/nips/GoldbergZRXN10}. We repeat each experiment five times, and report the average results.

\paragraph{Results} We measure the recovery performance on the feature matrix $\X$ by the relative error $\|\widehat\X-\X\|_\mathrm{F}/\|\X\|_\mathrm{F}$. The classification performance is measured by the Hamming loss. More specially, after we got $\widehat Z$, we set $\hat Y_{ij}=1$ if $\sigma(\hat Z_{ij})>0.5$ and $0$ otherwise. The recovery performance on $\Y$ is then measured by $\sum_{i,j}\L_{0/1}(Y_{ij},\hat Y_{ij})/(nl)$ where $\L_{0/1}(\cdot,\cdot)$ is the zero-one loss. The results are shown in Table~\ref{tbl:simulation}. Note that we have $72$ results in total. We present $10$ results here and put all others in Appendix. From the results, we can see that, when data satisfy our assumption, our proposed COCO with the term $\|\widehat\X-\X\|_\mathrm{F}/\|\X\|_\mathrm{F}$ in the optimization objective is always better at $\X$ recovery. For the $\Y$ recovery, our proposal is  always better than two baselines, i.e., Maxide and Mc. Occasionally ($13$ among all $72$ cases) it is comparable to COCO-0. This would be reasonable since the term $\|\proj_{\Omega_X}(\widehat\X-\X)\|_\mathrm{tr}$ put more emphasis on feature recovery, and does not aid label recovery much. Comparing Maxide and Mc, we find that both two algorithms have the same recovery results on $\X$, but Maxide performs much worse on $\Y$ than MC. This may due to the fact that when recovery $\Y$, Mc uses additional information on the structure of $[X;\Y]$ instead of using only the non-perfect recovered feature data.

\begin{figure*}[!t]
\centering
\begin{minipage}[h]{1.3in}
\centering
\includegraphics[width= 1.3in]{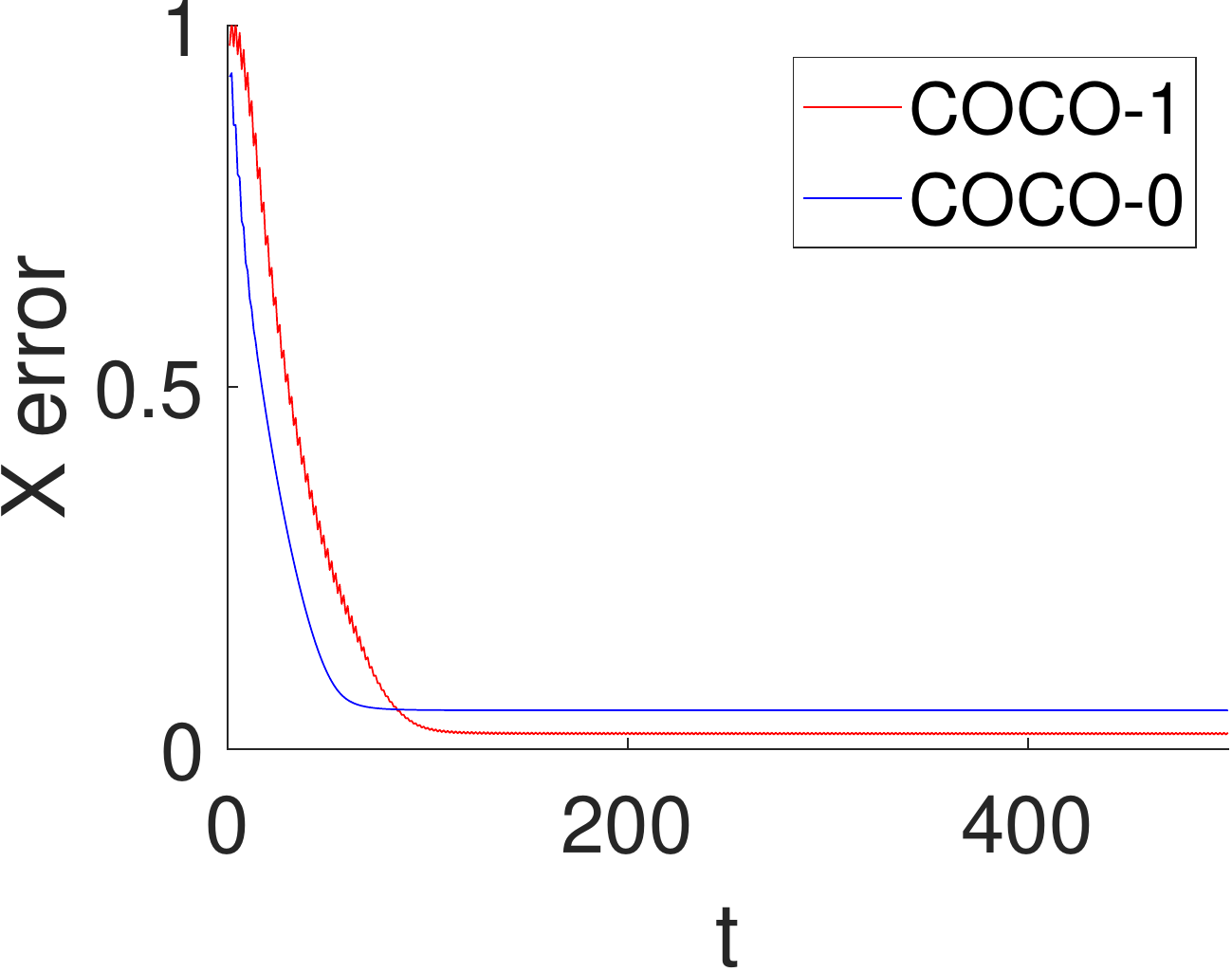}\\
\mbox{(a)}
\end{minipage}
\begin{minipage}[h]{1.3in}
\centering
\includegraphics[width= 1.3in]{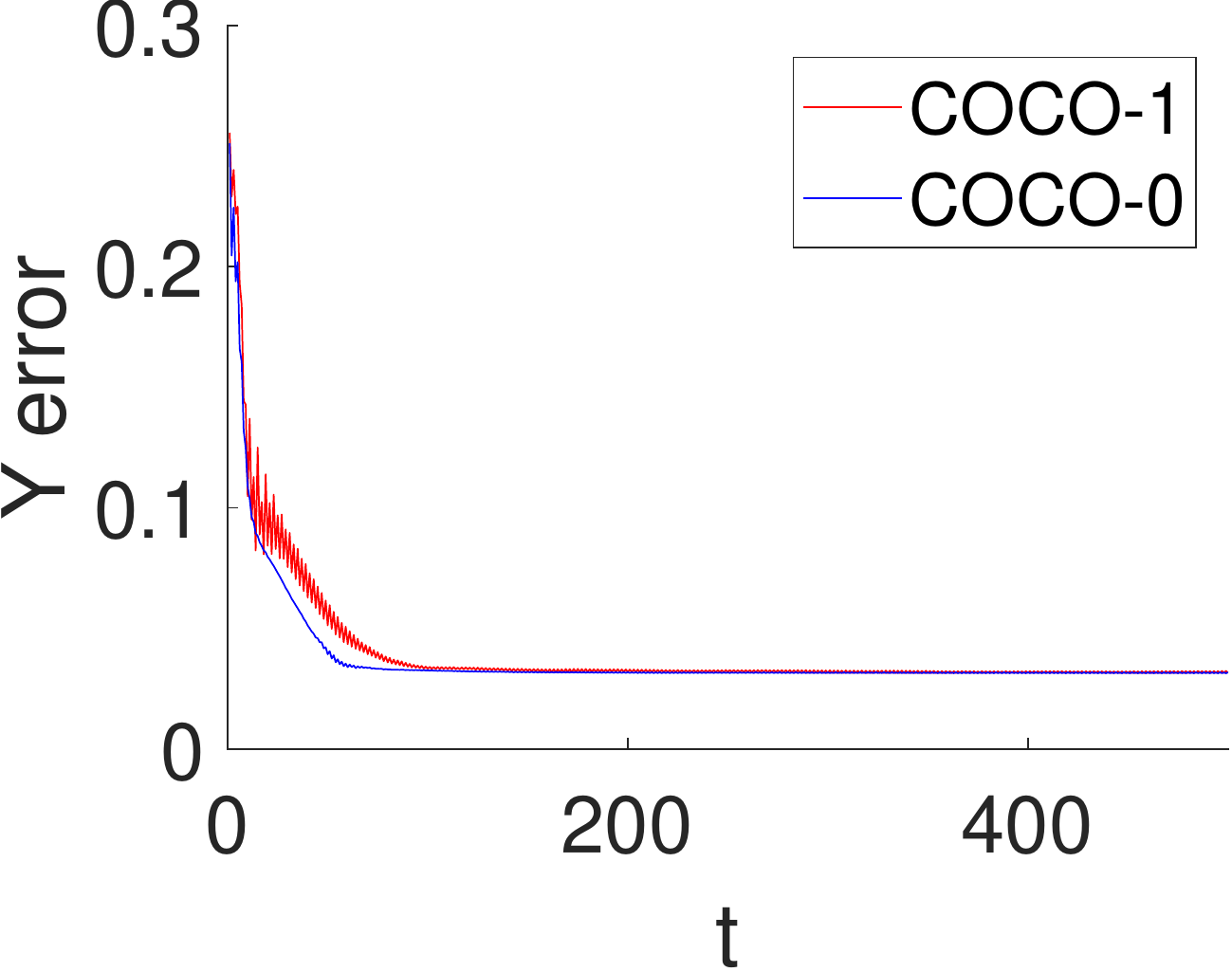}\\
\mbox{(b)}
\end{minipage}
\begin{minipage}[h]{1.3in}
\centering
\includegraphics[width= 1.3in ]{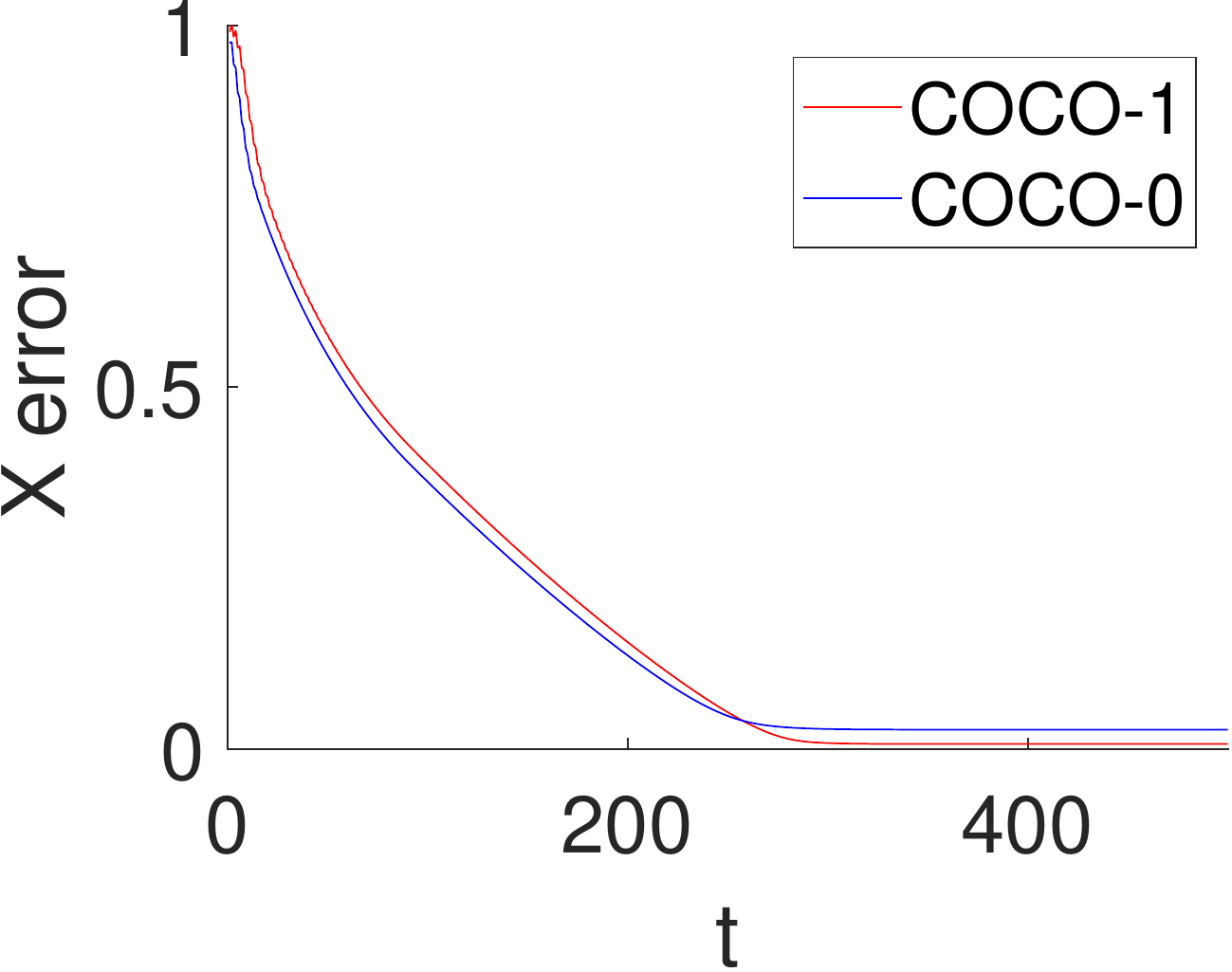}\\
\mbox{(c)}
\end{minipage}
\begin{minipage}[h]{1.3in}
\centering
\includegraphics[width= 1.3in ]{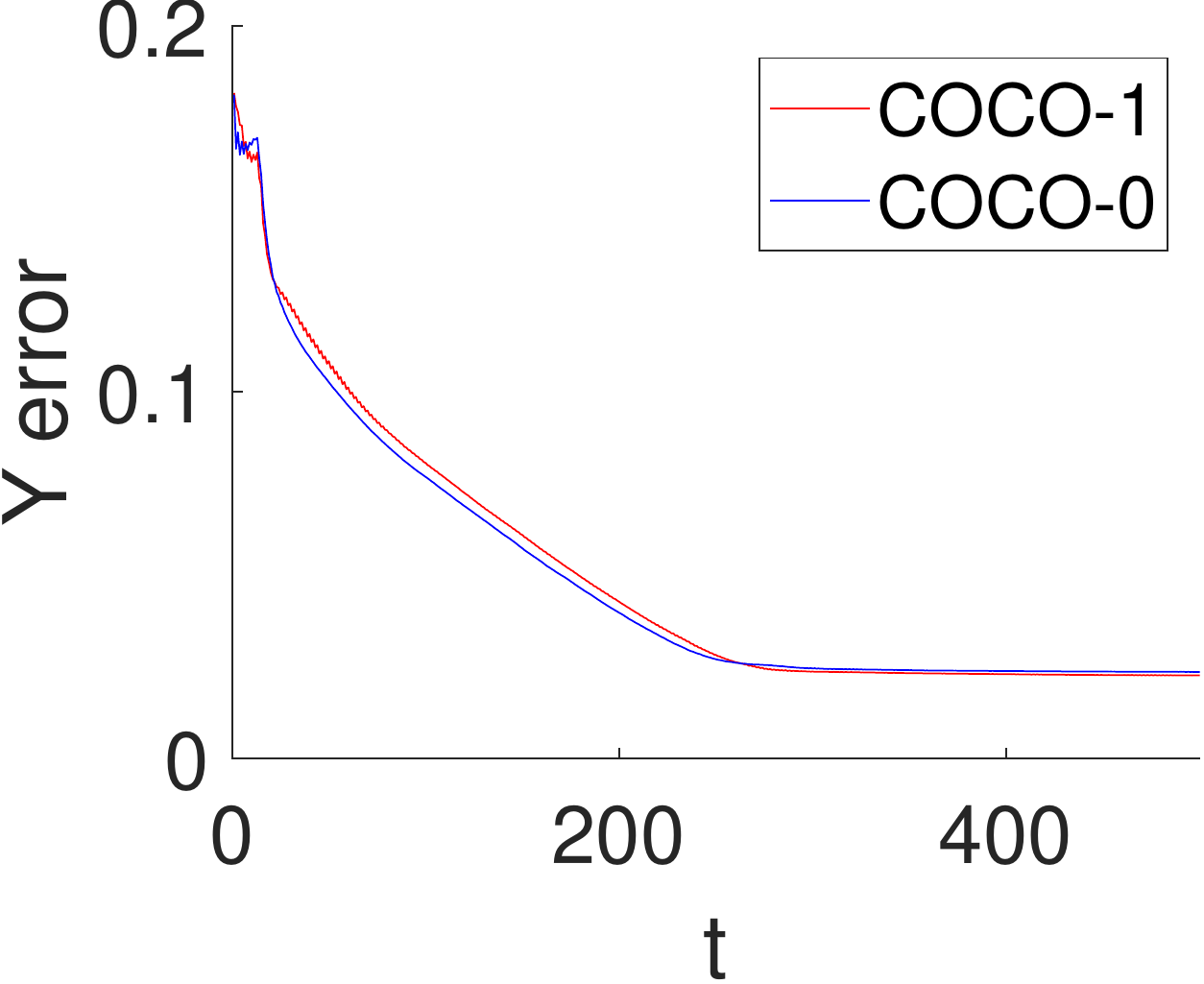}\\
\mbox{(d)}
\end{minipage}
\caption{\small Results of recovery errors when iteration number $t$ increases for the COCO-1 and COCO-0 algorithms. a) the recovery error of $\X$ for $n=1000, d=500,l=100,r=10,\omega=20$; b) the recovery error of $\Y$ for $n=1000, d=500,l=100,r=10,\omega=20$; c) the recovery error of $\X$ for $n=5000, d=500,l=100,r=20,\omega=20$; d) the recovery error of $\Y$ for $n=5000, d=500,l=100,r=20,\omega=20$.  }\label{fig:simulation}
\end{figure*}

To further study the impact of $\|\proj_{\Omega_X}(\widehat\X-\X)\|_\mathrm{tr}$ on the final performance, we illustrate how the recovery error of $\X$ and $\Y$ decrease when the iterations evolve in Figure~\ref{fig:simulation}. We can see that the $\X$ recovery error of COCO-1 decreases to a lower point when it converges, and get a slightly better recovery results than COCO-0. Although the $\Y$ recovery error also decreases to a lower point, the difference is not obvious. We can conclude that adding the term $\|\proj_{\Omega_X}(\widehat\X-\X)\|_\mathrm{tr}$ to the optimization objective can benefit $\X$ recovery.

\begin{table*}[!tp]
\caption{Recovery results (mean$\pm$std) over 5 trails on $10$ synthetic datasets. $n$ is the number of rows. $d$ is the number of features. $l$ is the number of labels. $r$ is the matrix rank and $\omega\%$ is the percentage of observed entries. COCO-1(0) is the proposed algorithm with $\lambda_2=1(0)$. The best result and comparable
ones (pairwise single-tailed t-tests at $95\%$ confidence level) in each row are bold. }\label{tbl:simulation}
\begin{center}
\scriptsize
\begin{tabular}{ccccccccc}
\hline
\multirow{2}{*}{$n$}
&\multirow{2}{*}{$d$}
&\multirow{2}{*}{$l$}
&\multirow{2}{*}{$r$}
&\multirow{2}{*}{$\omega$}
&\multicolumn{4}{c}{$\X$ recovery error}\\
\cline{6-9}
&&&&&COCO-1&COCO-0&Maxide&Mc\\
\hline
$1000$&$500$&$100$&$10$&$20$&$\mathbf{0.0221\pm0.0001}$&$0.0527\pm0.0004$&$0.5413\pm0.0027$&$0.5414\pm0.0027$\\
$1000$&$500$&$100$&$20$&$40$&$\mathbf{0.0148\pm0.0001}$&$0.0249\pm0.0001$&$0.3745\pm0.0026$&$0.3745\pm0.0026$\\
$1000$&$500$&$300$&$20$&$30$&$\mathbf{0.0188\pm0.0002}$&$0.0372\pm0.0004$&$0.5267\pm0.0013$&$0.5268\pm0.0013$\\
$1000$&$800$&$100$&$20$&$30$&$\mathbf{0.0137\pm0.0001}$&$0.0269\pm0.0003$&$0.5244\pm0.0013$&$0.5244\pm0.0013$\\
$5000$&$500$&$100$&$10$&$40$&$\mathbf{0.0082\pm0.0001}$&$0.0101\pm0.0001$&$0.4289\pm0.0019$&$0.4288\pm0.0019$\\
$5000$&$500$&$100$&$20$&$30$&$\mathbf{0.0061\pm0.0000}$&$0.0153\pm0.0000$&$0.6581\pm0.0011$&$0.6579\pm0.0011$\\
$5000$&$500$&$100$&$20$&$40$&$\mathbf{0.0064\pm0.0000}$&$0.0107\pm0.0000$&$0.5515\pm0.0011$&$0.5514\pm0.0011$\\
$5000$&$500$&$300$&$20$&$30$&$\mathbf{0.0066\pm0.0000}$&$0.0154\pm0.0000$&$0.6578\pm0.0012$&$0.6574\pm0.0012$\\
$10000$&$500$&$100$&$10$&$20$&$\mathbf{0.0049\pm0.0000}$&$0.0157\pm0.0001$&$0.7387\pm0.0008$&$0.7386\pm0.0008$\\
$10000$&$500$&$100$&$10$&$30$&$\mathbf{0.0054\pm0.0000}$&$0.0098\pm0.0001$&$0.6292\pm0.0011$&$0.6291\pm0.0011$\\
\hline
\multirow{2}{*}{$n$}
&\multirow{2}{*}{$d$}
&\multirow{2}{*}{$l$}
&\multirow{2}{*}{$r$}
&\multirow{2}{*}{$\omega$}
&\multicolumn{4}{c}{$\Y$ recovery error}\\
\cline{6-9}
&&&&&COCO-1&COCO-0&Maxide&Mc\\
\hline
$1000$&$500$&$100$&$10$&$20$&$\mathbf{0.0318\pm0.0008}$&$\mathbf{0.0319\pm0.0009}$&$0.2987\pm0.0023$&$0.2829\pm0.0018$\\
$1000$&$500$&$100$&$20$&$40$&$\mathbf{0.0221\pm0.0007}$&$\mathbf{0.0221\pm0.0007}$&$0.2560\pm0.0016$&$0.1937\pm0.0009$\\
$1000$&$500$&$300$&$20$&$30$&$\mathbf{0.0310\pm0.0003}$&$0.0314\pm0.0004$&$0.2697\pm0.0017$&$0.2392\pm0.0008$\\
$1000$&$800$&$100$&$20$&$30$&$\mathbf{0.0285\pm0.0006}$&$0.0283\pm0.0006$&$0.2644\pm0.0032$&$0.2659\pm0.0015$\\
$5000$&$500$&$100$&$10$&$40$&$\mathbf{0.0158\pm0.0004}$&$0.0161\pm0.0004$&$0.3269\pm0.0019$&$0.0827\pm0.0004$\\
$5000$&$500$&$100$&$20$&$30$&$\mathbf{0.0177\pm0.0003}$&$0.0181\pm0.0003$&$0.3205\pm0.0010$&$0.1355\pm0.0007$\\
$5000$&$500$&$100$&$20$&$40$&$\mathbf{0.0143\pm0.0001}$&$0.0146\pm0.0002$&$0.3196\pm0.0007$&$0.0935\pm0.0008$\\
$5000$&$500$&$300$&$20$&$30$&$\mathbf{0.0196\pm0.0001}$&$0.0203\pm0.0001$&$0.3204\pm0.0008$&$0.1317\pm0.0003$\\
$10000$&$500$&$100$&$10$&$20$&$\mathbf{0.0212\pm0.0003}$&$0.0220\pm0.0003$&$0.3389\pm0.0011$&$0.1360\pm0.0004$\\
$10000$&$500$&$100$&$10$&$30$&$\mathbf{0.0175\pm0.0003}$&$0.0181\pm0.0003$&$0.3429\pm0.0008$&$0.0896\pm0.0008$\\
\hline
\end{tabular}
\end{center}
\end{table*}


\subsection{Experimental Results on Real Data}

We evaluate the proposed algorithm on real data. Here we will evaluate the performance using the CAL500 dataset~\cite{DBLP:journals/taslp/TurnbullBTL08}. CAL500 is a music dataset containing $502$ instances, $68$ features, $174$ labels. As we previously shown in Figure~\ref{illustration-general}, CAL500's annotation matrix does not have the low-rank or approximately low-rank property. In this experiment, we will not only report the recovery performance of COCO, but also use the recovered data to train new classifiers, and report the test error of the trained classifier.

\paragraph{Settings and Baselines} We will first divide the datasets into two parts, $80\%$ for training and $20\%$ for testing. For the $80\%$ training data, we will randomly sample $20\%$ as observed data, and make all other entries unobserved. We will use the same parameter setting as Section~\ref{sec:expsim}, except that the step size will keep decaying without stopping. Here we will also compare with Maxide and Mc. For the two compared methods, we use the default parameter setting in their original codes. After the data are recovered, we use the state-of-the-art multi-label classification method called
LIMO (label-wise and Instance-wise margins optimization)~\cite{DBLP:conf/icml/WuZ17} and a single hidden layer neural network to test the generalization performance when using the recovered data to train a classifier. To make a fair comparison, we also use the clean data to train a classifier and record its test error, which can be counted as the best baseline for the current model. We will call this method the oracle. All the experiments are repeated twenty times and report the average results.

\paragraph{Results}
The results are reported in Table~\ref{tbl:cal500}. We can see that our proposed COCO achieves the best recovery results among all three methods. For the generalization performance, we can see that our method also achieves the best results in all comparable methods, and it is more closer to the baseline using clean data.

\begin{table*}[!t]
\caption{Experimental results on the CAL500 datasets. The recovery error of $\X$, $\Y$, as well as the test error when using the recovered data to train classifiers LIMO and NN are shown. The last line gives the test error of the oracle, i.e., using the clean data to train a classifier and report its test error. The best result and comparable ones (pairwise single-tailed t-tests at $95\%$ confidence level, except Oracle) in each column are bold.}\label{tbl:cal500}
\begin{center}
\begin{tabular}{cccccc}
\hline
\multirow{2}{*}{}
&\multicolumn{2}{c}{Recovery Error}&\multirow{2}{*}{}
&\multicolumn{2}{c}{Test Error}
\\\cline{2-3}\cline{5-6}
&X-error&Y-error&&LIMO&NN\\
\hline
COCO&  $\mathbf{0.6477\pm0.0336}$&$\mathbf{0.1025\pm0.0004}$&&$\mathbf{0.1513\pm0.0068}$&$\mathbf{0.1469\pm0.0041}$\\
Maxide&$0.8588\pm0.0170$&$0.1074\pm0.0009$&&$0.1602\pm0.0081$&$0.1498\pm0.0032$\\
Mc&    $0.8608\pm0.0168$&$0.1430\pm0.0008$&&$0.1533\pm0.0054$&$0.1486\pm0.0036$\\
\hline
Oracle&$0.0000\pm0.0000$&$0.0000\pm0.0000$&&$0.1388\pm0.0021$&$0.1369\pm0.0020$\\
\hline
\end{tabular}
\end{center}
\end{table*}


\section{Conclusion}~\label{sec:con}
In this paper, we considered the problem where both features and labels have missing values in weakly supervised multi-label learning. Realizing that previous methods either recover the features ignoring supervised information, or make unrealistic assumptions, we proposed a new method to deal with such problems. More specifically, we considered a latent matrix generating the label matrix, and considering labels are correlated, such a latent matrix together with features form a big low-rank matrix. We then gave our optimization objective and algorithm motivated by the elastic net. Experimental results on both simulated and real-world data validated the effectiveness of our proposed methods.

\subsubsection*{Acknowledgments} We want to thank Bo-Jian Hou for discussion and polishing of the paper.

\bibliography{ref}
\bibliographystyle{unsrt}

\end{document}